\documentclass{article}
\pdfoutput=1
\usepackage{microtype}
\usepackage{graphicx}
\usepackage{subfigure}
\usepackage{booktabs} 
\usepackage{hyperref}
\usepackage{amsmath,amssymb}

\usepackage[accepted]{icml2018}
\usepackage[title]{appendix}
\icmltitlerunning{Continual Reinforcement Learning with Complex Synapses}

\newcommand{\beginsupplement}{%
        \setcounter{table}{0}
        \renewcommand{\thetable}{S\arabic{table}}%
        \setcounter{figure}{0}
        \renewcommand{\thefigure}{S\arabic{figure}}%
     }

\makeatletter
\let\@fnsymbol\@arabic
\makeatother
\date{}

\begin{document}

\twocolumn[
\icmltitle{Continual Reinforcement Learning with Complex Synapses}

\icmlsetsymbol{equal}{*}

\begin{icmlauthorlist}
\icmlauthor{Christos Kaplanis}{impdoc,impbio}
\icmlauthor{Murray Shanahan}{impdoc,deep}
\icmlauthor{Claudia Clopath}{impbio}
\end{icmlauthorlist}

\icmlaffiliation{impdoc}{Department of Computing, Imperial College London}
\icmlaffiliation{deep}{Google DeepMind, London}
\icmlaffiliation{impbio}{Department of Bioengineering, Imperial College London}

\icmlcorrespondingauthor{Christos Kaplanis}{christos.kaplanis14@imperial.ac.uk}

\icmlkeywords{Machine Learning, ICML}
\vskip 0.3in
]

\printAffiliationsAndNotice{} 
\pagenumbering{gobble}
\begin{abstract}
Unlike humans, who are capable of \textit{continual learning} over their lifetimes, artificial neural networks have long been known to suffer from a phenomenon known as \textit{catastrophic forgetting}, whereby new learning can lead to abrupt erasure of previously acquired knowledge. Whereas in a neural network the parameters are typically modelled as scalar values, an individual synapse in the brain comprises a complex network of interacting biochemical components that evolve at different timescales. In this paper, we show that by equipping tabular and deep reinforcement learning agents with a synaptic model that incorporates this biological complexity \cite{benna2016computational}, catastrophic forgetting can be mitigated at \textit{multiple} timescales. In particular, we find that as well as enabling continual learning across sequential training of two simple tasks, it can also be used to overcome \textit{within}-task forgetting by reducing the need for an experience replay database.
\end{abstract}

\section{Introduction}
\label{intro}
One of the outstanding enigmas in computational neuroscience is how the brain is capable of \textit{continual} or \textit{lifelong learning} \cite{wixted2004psychology}, acquiring new memories and skills very quickly while robustly preserving old ones. Synaptic plasticity, the ability of the connections between neurons to change their strength over time, is widely considered to be the physical basis of learning in the brain and knowledge is thought to be \textit{distributed} across neuronal networks, with individual synapses participating in the storage of several memories. Given this overlapping nature of memory storage, it would seem that synapses need to be both labile in response to new experiences and stable enough to retain old memories - a paradox often referred to as the \textit{stability-plasticity dilemma} \cite{carpenter1987art}.

Artificial neural networks also have a distributed memory but, unlike the brain, are prone to \textit{catastrophic forgetting} \cite{mccloskey1989catastrophic,french1999catastrophic}; when trained on a nonstationary data distribution, such as two distinct tasks in sequence, a network can quickly forget what it learnt from earlier data. In reinforcement learning (RL), where data is typically accumulated online as the agent interacts with the environment, the distribution of experiences is often nonstationary over the training of a \textit{single} task, as well as across tasks, since (i) experiences are correlated in time and (ii) the agent's policy changes as it learns. A typical way of addressing nonstationarity of data in deep RL is to store experiences in a replay database and use it to \textit{interleave} old data and new data during training \cite{mnih2015human}. However, this solution does not scale well computationally as the number of tasks grows and the old data might also become unavailable at some point. Furthermore, it does not explain how the brain achieves continual learning, since the question remains as to how an ever-growing dataset is then stored without catastrophic forgetting.

One potential answer may arise from the experimental observations that synaptic plasticity occurs at a range of different timescales, including short-term plasticity \cite{zucker2002short}, long-term plasticity \cite{bliss1973long} and synaptic consolidation \cite{clopath2008tag}. Intuitively, the slow components to plasticity could ensure that a synapse retains memory of a long history of its modifications, while the fast components render the synapse highly adaptable to the formation of new memories, perhaps providing a solution the stability-plasticity dilemma.

In this paper, we explore whether a biologically plausible synaptic model \cite{benna2016computational}, which abstractly models plasticity over a range of timescales, can be applied to mitigate catastrophic forgetting in a reinforcement learning context. Our work is intended as a proof of principle for how the incorporation of biological complexity to an agent's parameters can be useful in tackling the lifelong learning problem. By running experiments with both tabular and deep RL agents, we find that the model helps continual learning across two simple tasks as well as within a single task, by allaying the necessity of an experience replay database, indicating that the incorporation of different timescales of \textit{plasticity} can correspondingly result in improved \textit{behavioural} memory over distinct timescales. Furthermore, this is achieved even though the process of synaptic consolidation has no prior knowledge of the timing of changes in the data distribution. 

\section{Background}
\label{background}
\subsection{The Benna-Fusi Model}
In this paper, we make use of a synaptic model that was originally derived to maximise the expected signal to noise ratio (SNR) of memories over time in a population of synapses undergoing continual plasticity in the form of random, uncorrelated modifications \cite{benna2016computational}. The model assumes that a synaptic weight $w$ at time $t$ is determined by its history of modifications up until that time $\Delta w(t')$, which are filtered by some kernel $r(t-t')$, such that
\begin{equation} \label{eq:modifications}
w(t)=\sum_{t'<t}{\Delta w(t')r(t-t')}.
\end{equation}
While constraining the variance of the synaptic weights to be finite, the expected (doubly logarithmic) area under the SNR vs. time curve of a given memory is typically maximised when $r(t)\sim t^{-\frac{1}{2}}$, i.e. the kernel decays with a power law. 

\begin{figure}[ht]
\begin{center}
\centerline{\includegraphics[width=\columnwidth]{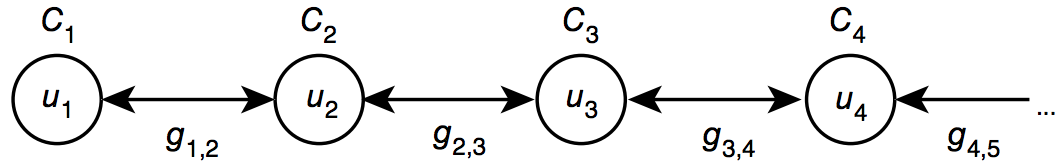}}
\centerline{\includegraphics[width=\columnwidth]{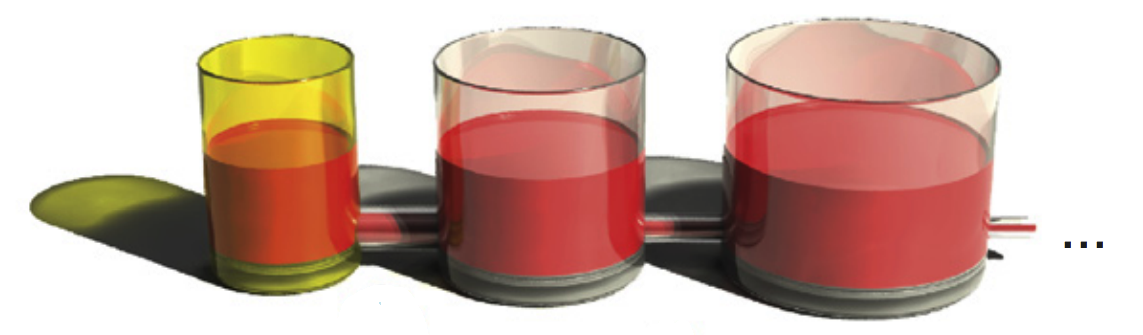}}
\caption{Diagrams adapted from \cite{benna2016computational} depicting the chain model (top) and the analogy to liquid flowing between a series of beakers of increasing size and decreasing tube widths (bottom).}
\label{benna_fusi_beakers}
\end{center}
\vskip -0.4in
\end{figure}

Implementing this model directly is impractical and unrealistic, since it would require recording the time and size of every synaptic modification; however, the authors show that the power law decay can be closely approximated by a synaptic model consisting of a finite chain of $N$ communicating dynamic variables (as depicted in Figure \ref{benna_fusi_beakers}). The dynamics of each variable $u_k$ in the chain are determined by interaction with its neighbours in the chain: 
\begin{equation}
C_k \frac{du_k}{dt}=g_{k-1,k}(u_{k-1}-u_k)+g_{k,k+1}(u_{k+1}-u_k)
\end{equation}
except for $k=1$, for which we have
\begin{equation}\label{eq:u1}
C_1 \frac{du_1}{dt}=\frac{dw_{ext}}{dt}+g_{1,2}(u_2-u_1)
\end{equation}
where $\frac{dw_{ext}}{dt}$ corresponds to a continuous form of the $\Delta w(t')$ updates (Equation \ref{eq:modifications}). For $k=N$, there is a leak term, which is constructed by setting $u_{N+1}$ to $0$. The synaptic weight itself $w$ is just read off from the value of $u_1$, while the other variables are hidden and have the effect of regularising the value of the weight by the history of its modifications. 

From a mechanical perspective, one can draw a comparison between the dynamics of the chain of variables and liquid flowing through a series of beakers with different base areas $C_k$ connected by tubes of widths $g_{k-1,k}$ and $g_{k,k+1}$, where the value of a $u_k$ variable corresponds to the level of liquid in the beaker (Figure \ref{benna_fusi_beakers}).

Given a finite number of beakers per synapse, the best approximation to a power law decay is achieved by exponentially increasing the base areas of the beakers and exponentially decreasing the tube widths as you move down the chain, such that $C_k = 2^{k-1}$ and $g_{k,k+1} \propto 2^{-k-2}$. Beakers with wide bases and connected by smaller tubes will necessarily evolve at longer timescales. From a biological perspective, the dynamic variables can be likened to reversible biochemical processes that are related to plasticity and occur at a large range of timescales.

Importantly, the model abstracts away from the causes of the synaptic modifications $\Delta w$ and so is amenable for testing in different learning settings. In the original paper \cite{benna2016computational}, the model was shown to extend lifetimes of random, uncorrelated memories in a perceptron and a Hopfield network, while in this work we test the capacity of the model to mitigate behavioural forgetting in more realistic tasks where synaptic updates are unlikely to be uncorrelated.

In all our experiments, we simulated the Benna-Fusi ODEs using the Euler method for numerical integration. 

\subsection{Reinforcement Learning}
All experiments in this paper were conducted in an RL paradigm. The RL setting is formalised as a Markov Decision Process (MDP) defined by a tuple $\langle \mathcal{S},\mathcal{A},p_s,r \rangle$, whereby at time step $t$, an agent observes the state $s_t \in \mathcal{S}$, takes an action $a_t \in \mathcal{A}$, resulting in a reward $r(s_t,a_t)$ and transition to the next state $s_{t+1}$ with probability $p_s(s_{t+1}| s_t,a_t)$. The goal of the agent is to find a policy, defined by a probability distribution over actions given the state $\pi (a_t | s_t)$, that maximises its expected sum of discounted future rewards:
\begin{equation}
\pi^* = \arg \max_{\pi} \sum_t \mathbb{E}_{\pi} [ r(s_t,a_t) ]
\end{equation}
where $\mathbb{E}_{\pi}$ is the expectation under the reward distribution defined by policy $\pi$.
\subsubsection{Q-learning}
Q-learning \cite{watkins1992q} is a well-known reinforcement learning algorithm that involves learning the Q-values for each state-action pair which, given a policy $\pi$, are defined as:
\begin{equation}
Q^{\pi}(s,a) = \mathbb{E}_{\pi} \big[ \sum_{i=t}^\infty \gamma^{i-t} r_{i} | s_t = s, a_t = a \big]
\end{equation}
where $\gamma$ is a temporal discount factor. By sampling experiences in the form $(s_t,a_t,r_t,s_{t+1})$ from a sufficiently exploratory policy and using them to update the Q-values as follows:
\begin{align}
\delta_t & \leftarrow r_t + \gamma V(s_{t+1}) - Q(s_t,a_t) \\
Q(s_t,a_t) & \leftarrow Q(s_t,a_t)+\eta \delta_t,
\end{align}
where $\eta$ is the learning rate and $V(s_{t+1})=\max_a Q(s_{t+1},a)$, $Q$ will eventually converge to $Q^*$, the value function of the optimal policy $\pi^*$, which is derived as 
\begin{equation}
\pi^*(a|s) = \begin{cases}
		1, & \text{if $a=\arg \underset{a'}{\max}\, Q^*(s,a')$},\\
        0, & \text{otherwise.}
        	  \end{cases}
\end{equation}
A common policy used for training, and the one used in this paper, is $\epsilon$-greedy, whereby with probability $1-\epsilon$ the agent chooses the action with the highest Q-value and, with probability $\epsilon$, chooses an action uniformly at random.

In this paper, we use a variant of Q-learning called `naive' $Q(\lambda)$ \cite{sutton1998reinforcement}, which can speed up the convergence of Q-learning and involves maintaining an eligibility trace $e(s,a)$ for each state-action pair. At each time step, all eligibility traces are updated as follows:
\begin{equation}
e_{t}(s,a) = \begin{cases}
		1, & \text{if $s_t=s$ and $a_t=a$}, \\
        \gamma \lambda e_{t-1}(s,a), & \text{otherwise.}
        	\end{cases}
\end{equation}
where $\lambda \in [0,1]$ is a constant decay parameter. All Q-values are then updated by:
\begin{equation}
Q(s,a) \leftarrow Q(s,a) + \eta \delta e(s,a)
\end{equation}
As is elucidated further on in the Experiments section, the eligibility traces can also be used to modulate the rate of synaptic consolidation to improve memory retention.

\subsubsection{Deep Q Networks}
In high-dimensional, continuous state spaces, it is infeasible to maintain a table of Q-values for all state-action pairs; in order to learn a good policy, an agent must be able to use its experience to generalise to previously unseen situations. Deep Q Networks (DQN) \cite{mnih2015human} are artificial neural networks that are trained to approximate a mapping from states to Q-values by optimising the following cost function:
\begin{equation}
L(\theta) = \mathbb{E}_{(s,a,r,s')\sim D}\left[\left(r + \gamma V(s';\theta^-) - Q(s,a;\theta)\right)^2 \right]
\end{equation}
where $\theta$ are the parameters of the network, $V(s';\theta^-)=max_{a'}Q(s,a';\theta^-)$ and $\theta^-$ are the parameters of an older version of the network (referred to as the target network) used to counteract instability in training due to quickly changing target values. $D$ is the \textit{experience replay database}, which records the agent's experiences in a FIFO queue and is sampled from at random during training. Consecutive experiences are usually highly correlated with one another and thus training in an online fashion can cause the network to \textit{overfit} to recent data; by jumbling together old and new data, the database thus plays an essential role in \textit{decorrelating} updates to the network and preventing catastrophic forgetting of older experiences. In some of our experiments described later on, we show that equipping the parameters of the network with the Benna-Fusi model can attenuate the need for an experience replay database due to better retention of older memories.

In our experiments, we trained the agents using a \textit{soft Q-learning} objective \cite{haarnoja2017reinforcement}, which generalises Q-learning by simultaneously maximising the entropy of the agent's policy:
\begin{equation}
\pi^* = \arg \max_{\pi} \sum_t \mathbb{E}_{\pi} [ r(s_t,a_t) + \alpha \mathcal{H}(\pi (\cdot | s_t))]
\end{equation}
where $\alpha$ is a constant that controls the balance between reward and entropy maximisation.

One benefit of soft Q-learning is that it can generate a more robust policy as it encourages the agent to learn multiple solutions to the task and, in our experiments with DQN, we found that it helped to stabilise performance over time.

\section{Experiments}
The overarching goal of the experiments was to test whether applying the Benna-Fusi model to an agent's parameters could enhance its ability to learn continually in an RL setting. Our aim was to demonstrate the \textit{potential} for the model in enabling continual learning and, for this reason, we tested it in relatively simple settings, where catastrophic forgetting is nevertheless still an issue.

The first experiments, which apply the model in a simple tabular Q-learning agent, were intended to serve as a proof of principle and as a means to gaining an intuition of the mechanics of the model through visualisation. Subsequently, we tested it in a deep RL agent to evaluate its effect on the agent's ability to learn continually across two simple tasks and also within a single task. The only Benna-Fusi parameters that were varied across experiments were the first tube width ($g_{1,2}$) and the number of hidden variables, which jointly determine the range of timescales that the model can capture.

\subsection{Continual Q-learning}
The first set of experiments were conducted in order to test whether applying the Benna-Fusi model to tabular Q-values could be used to facilitate continual reinforcement learning in a simple grid-world setting.
\subsubsection{Experimental Setup}
The environment consisted of 100 states organised into a 10x10 two-dimensional grid and the agent was equipped with 5 actions, 4 of which deterministically move the agent to a vertically or horizontally adjacent state and the last of which is a pick-up action that must be chosen to collect the reward when in the correct location. The agent was trained alternately on two different tasks; in the first, the reward was located in the upper right-hand corner of the grid and, in the second, it was in the bottom left-hand corner. An episode was terminated if the agent reached the goal state and successfully picked up the reward, or if it took a maximum number of steps without reaching the goal. In order to test the agent's ability to learn continually, the goal location was switched every 10,000 episodes (one epoch) and the time taken for the agent to \textit{relearn} to capture the reward was measured.

Three different agents were trained and compared:
\begin{itemize}
\item A control agent trained in an online fashion with naive $Q(\lambda)$ using an $\epsilon$-greedy policy.
\item A Benna-Fusi agent, also trained with naive $Q(\lambda)$, but for which the tabular Q-values were modelled as a Benna-Fusi synapses, each with their own chain of interacting dynamic variables. For a given state-action pair $(s,a)$, we denote the first variable in the chain as $Q^1(s,a)$, which corresponds to $u_1$ in Equation \ref{eq:u1} and is the `visible' Q-value that determines the agent's policy at any time. The Q-learning updates $\eta \delta(t) e_t(s,a)$ correspond to the $\Delta w(t)$ modifications. The deeper variables in the chain $Q^k(s,a)$, with $k>1$, can be thought of as `hidden' Q-values that `remember' what the visible Q-value function was over longer timescales and regularise it by its history.
\item A modified Benna-Fusi agent, whereby at every time step, the flow from $Q^k(s,a)$ to $Q^{k+1}(s,a)$ for all variables in the chain was scaled by a multiple of the eligibility trace $e_t(s,a)$. The flow from shallow variables to deeper variables in the chain can be thought of as a process of \textit{consolidation} of the synapse, or in this case Q-value. The rationale for modulating this flow by the eligibility trace is that it only makes sense to consolidate parameters that are actually being used and modified; for example, if a state $s$ has not been visited for a long time, we should not become increasingly sure of any of the Q-values $Q^1(s,a)$.
\end{itemize}
In a Benna-Fusi chain of length $N$, $\frac{C_1}{g_{1,2}}$ and $\frac{C_N}{g_{N,N+1}}$ determine the shortest and longest memory timescales of the hidden variables respectively. In our experiments, we set $g_{1,2}$ to $10^{-5}$ to correspond roughly to the minimum number of Q-learning updates per epoch, and the number of variables in each chain to 3, all of which were initialised to 0. The ODEs were numerically integrated after every Q-learning update with a time step of $\Delta t=1$. A table of all parameters used for simulation is shown in Table S1.

\subsubsection{Results}
\begin{figure}[!ht]
\begin{center}
\centerline{\includegraphics[width=\columnwidth]{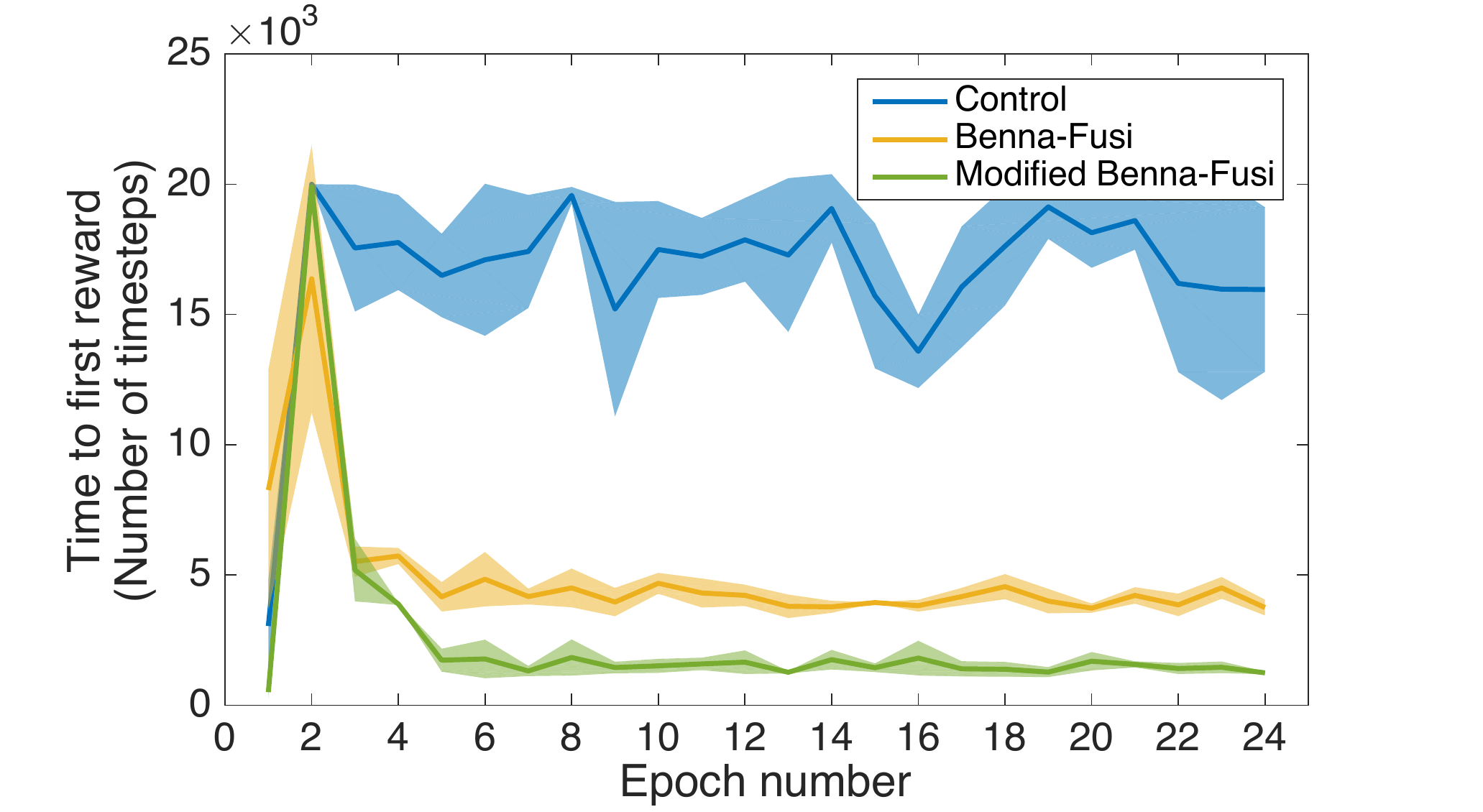}}
\centerline{\includegraphics[width=\columnwidth]{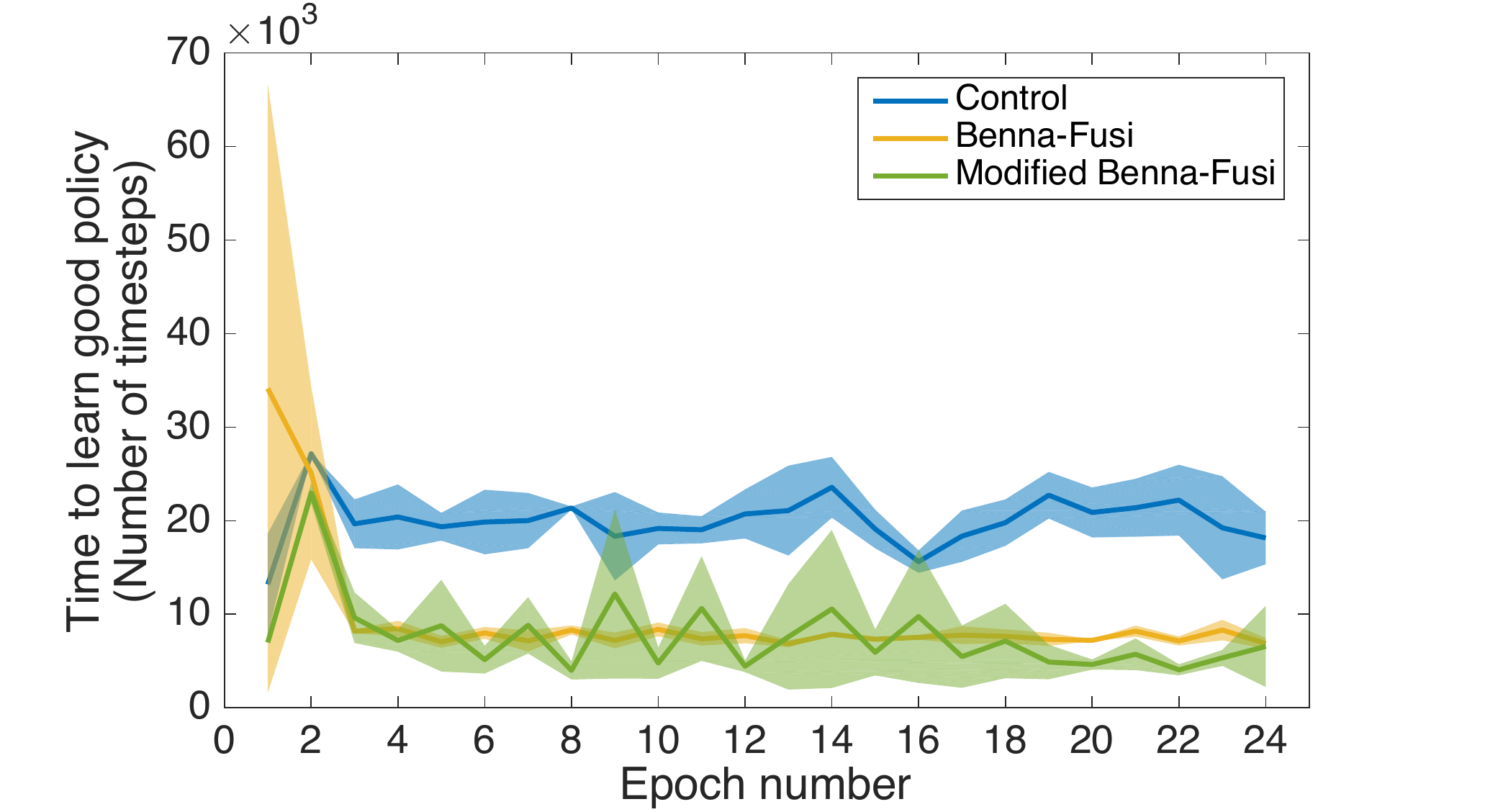}}
\caption{(Top) How long it took each agent to relearn to navigate to the first reward at the beginning of each epoch. (Bottom) How many time steps it took for the 20-episode moving average of episode lengths to drop below 13, as a measure of how long it took to (re)learn a good policy. Mean over 3 runs with 1 s.d. error bars.}
\label{trial_length_comparison}
\end{center}
\vskip -0.4in
\end{figure}

The Benna-Fusi agents learned to switch between good policies for each task significantly faster than the control agent, with the modified Benna-Fusi agent being the quickest to relocate the reward for the first time at the beginning of each epoch (Figure \ref{trial_length_comparison}). After the agents have learned to perform the task in the first epoch, it takes them all a long time to find the reward when its location is switched to the opposite corner at the beginning of the second epoch, since their policies are initially tuned to move actively away from the reward. After subsequent reward switches, however, while the control agent continues to take a long time to relearn a good policy due to the negative transfer between the two tasks, both Benna-Fusi agents learn to re-attain a good level of performance on the task much faster than after the initial task-switch (see bottom of Figure \ref{trial_length_comparison}).

\begin{figure*}[!ht]
\begin{center}
\includegraphics[width=\textwidth]{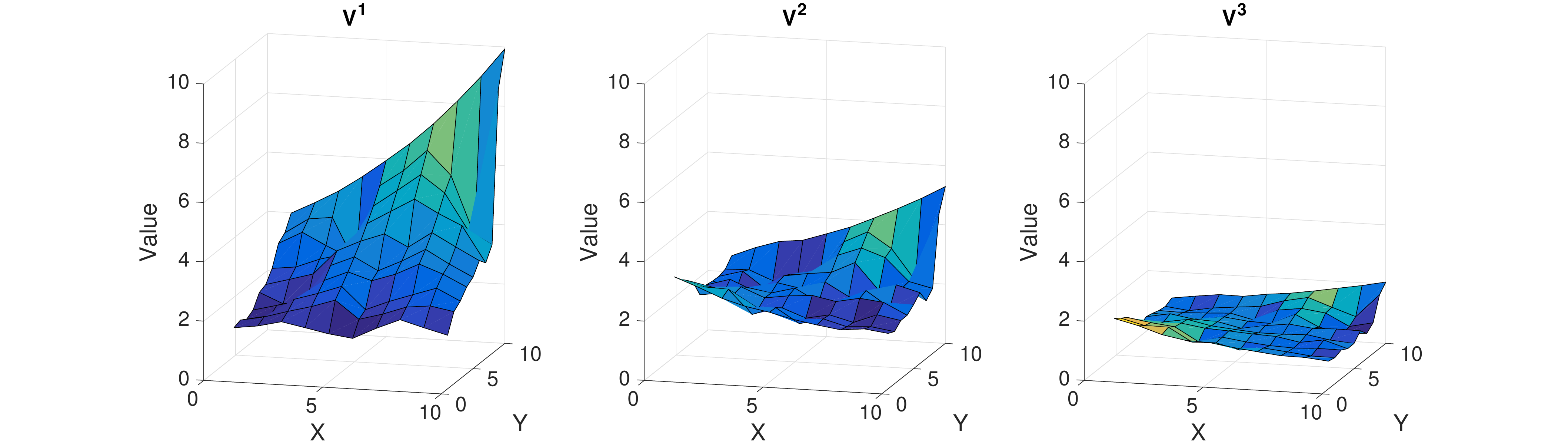}
\caption{Surface plots of a snapshot of the visible ($V^1$) and hidden ($V^2$ and $V^3$) values of each state during training.  While $V^1$ only appears to retain information about the current reward at (10,10), $V^2$ and $V^3$ still remember that there is value at (0,0). When the reward location is switched back to (0,0), flow from the deeper variables in the chain back into $V^1$ make it easier for the agent to recall the previous reward location. See \url{https://youtu.be/_KgGpT-sjAU} for animation of values over training.}
\label{hidden_values}
\end{center}
\end{figure*}

In order to visualise the role of the hidden variables of the Benna-Fusi model in enabling continual learning, we define $V^k(s):= \max_{a'}Q^k(s,a')$; for $k=1$, this simply corresponds to the traditional value function $V(s)$ and, for $k>1$, one can interpret $V^k(s)$ as a `hidden' value function that records the value function over longer timescales. Figure \ref{hidden_values} shows a snapshot of the $V^k$ values during training, depicting how the deeper Benna-Fusi variables remember the Q-values at a longer timescale, enabling the agent to quickly recall the location of the previous reward. See \url{https://youtu.be/_KgGpT-sjAU} for an animation of this process.

\subsection{Continual Multi-task Deep RL}
The next set of experiments were to test if we could observe similar benefits to continual learning if the Benna-Fusi model were applied to the parameters of a deep RL agent alternately performing two simple tasks. While having a better memory retention for tabular Q-values has a direct impact on an agent's ability to recall a previous policy, it is less obvious that longer memory lifetimes in individual synapses should yield better behavioural memory in a distributed system such as a deep Q-network.

\subsubsection{Experimental Setup}
The two tasks used for this experiment were Cart-Pole\footnote{The version used was CartPole-v1 from the OpenAI Gym \cite{brockman2016openai}} and Catcher, which were suitable for training on the same network as they have the same size of state and action spaces respectively.

Similarly to the tabular Q-learning experiments, an agent was trained alternately on the two tasks (for 40 epochs of 20,000 episodes) and, as a measure of its ability to learn continually, the time taken for the agent to (re)learn the task after every switch was recorded. A task was deemed to have been (re)learnt if a moving average of the reward per episode moved above a predetermined level (450 for Cart-Pole, which has max reward 500, and 10 for Catcher, which has max reward about 14).

Experiments were run with two types of agent, a control agent and a Benna-Fusi agent. In order to ensure that the difference in performance of the two agents was not just due to differences in the effective learning rate (which is likely to be lower in the Benna-Fusi agent as the parameters are regularised by the hidden variables), the control agent was run with several different learning rates. The Benna-Fusi agent was only run with $\eta=0.001$.

The control agent was essentially a DQN \cite{mnih2015human} with two fully connected hidden layers of 400 and 200 ReLUs respectively, but with a number of modifications that were made in order to give it as good a chance as possible to learn continually. The network was trained with the soft Q-learning objective \cite{haarnoja2017reinforcement}, which helped to stabilise learning in each task, presumably by maintaining a more diverse set of experiences in the replay database\footnote{In particular, we found this more effective than having a larger replay database, decaying $\epsilon$ to a positive value or just having a softmax policy.}. Furthermore, as in \cite{kirkpatrick2017overcoming}, while the network weights were shared between tasks, each layer of the network was allowed to utilise task-specific gains and biases, such that computations at each layer were of the form:
\begin{equation}
y_i = g_i^c\left(b_i^c+\sum_j W_{ij}x_j \right)
\end{equation}
where $c$ indexes the task being trained on. This helped overcome the issue of training a network on two different Q-functions, which has been reported to be very challenging even as a regression task \cite{rusu2015policy}.

The experience replay database had a size of 2000, from which 64 experiences were sampled for training with Adam \cite{kingma2014adam} at the end of every episode. Crucially, the database was cleared at the end of every epoch in order to ensure that the agent was only training on one task at a time. The agent was $\epsilon$-greedy with respect to the stochastic soft Q-learning policy and $\epsilon$ was decayed from $1$ to almost $0$ over the course of each epoch. Finally, `soft' target network updates were used as in \cite{lillicrap2015continuous}, rather than hard periodic updates used in the original DQN. 

The Benna-Fusi agent was identical to the control agent, except that each network parameter was modelled as a Benna-Fusi synapse with 30 variables with $g_{1,2}$ set to $0.001625$, ensuring that the longest timescale ($\propto \frac{C_{30}}{g_{30,31}}$) comfortably exceeded the total number of updates over training ($\approx 2^{25}$). In order to speed up computation, rather than simulate 64 time steps of the ODEs after every replay batch, these were approximated by conducting one Euler update with $\Delta t = 64$. For this reason, the effective flow between $u_1$ and $u_2$ was $64*0.001625=0.1$; if it were larger than 1 this would lead to instability or unwanted oscillations or negative $u$-values, so we could not increase $g_{1,2}$ much more. The complexity of the algorithm is $\mathcal{O}(mN)$, where $N$ is the number of trainable parameters in the network and $m$ is the number of Benna-Fusi variables per parameter. The compression of 64 Benna-Fusi updates into one resulted in the overall runtime being only 1.5-2 times longer than the control model.

The initial values of the hidden variables were normally distributed with variances decaying linearly with the depth in the chain, approximately matching the equilibrium distribution shown for random, uncorrelated memories in the original paper \cite{benna2016computational}. Furthermore, we incrementally allowed flow to occur from the deeper variables to the shallow ones so that the parameters were not constrained much by the random initialisation and only by hidden variables that have had enough time to adapt to the actual experiences of the agent. Specifically, flow from $u_{k+1}$ to $u_k$ was only enabled after $\frac{2^k}{g_{1,2}}$ gradient updates. 

A full table of parameters used can be seen in Table S2.

\subsubsection{Results}
\begin{figure}[ht]

\begin{center}
\centerline{\includegraphics[width=\columnwidth]{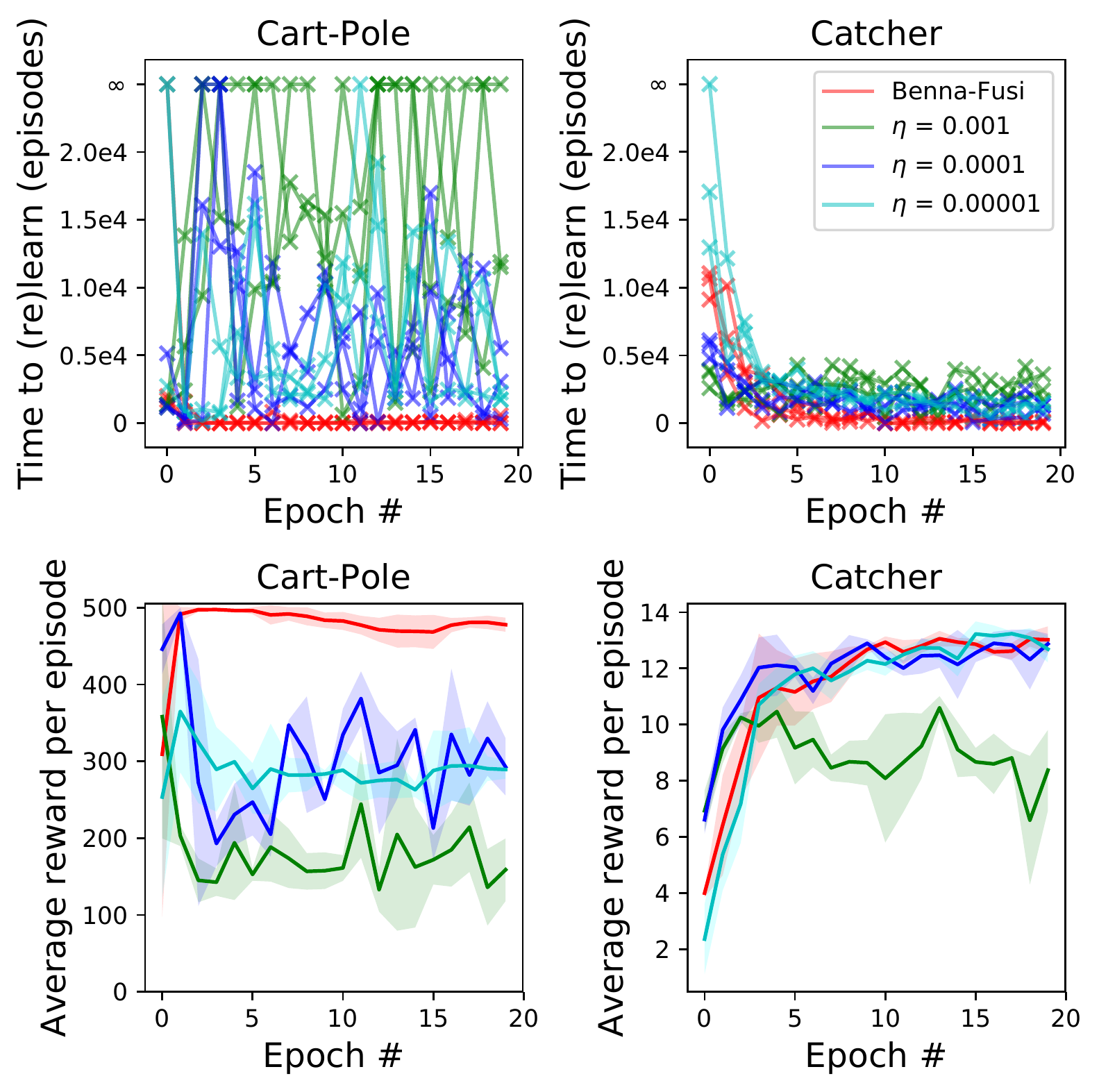}}

\caption{(Top) How long it took for the agents to relearn each task from the beginning of each epoch; the \# of training episodes needed for the 10 test-episode moving average of reward to surpass the threshold is plotted for 3 runs per agent. Runs that did not relearn within the are marked at $\infty$. (Bottom) Reward per episode averaged over each epoch for each task; means with s.d. error bars over 3 runs.}
\label{multitask_plots}
\end{center}
\vskip -0.4in
\end{figure}

Over the course of training the Benna-Fusi agent became faster at reaching at adequate level of performance on each task than the control agents (Figure \ref{multitask_plots}, top), thus demonstrating a better ability for continual learning. Interestingly, while the control agents were all able to learn Cart-Pole at the beginning of training, subsequent training on Catcher then left the network at a starting point that made it very hard or impossible for the agents to relearn Cart-Pole (as evidenced by the number of epochs where an adequate performance was never reached), exhibiting a severe case of catastrophic forgetting. The Benna-Fusi agent did not display this behaviour and, instead, relearned the task quickly in all epochs. It is important to note that parameters were chosen such that the control agents were all capable of learning a very good policy for either task when trained from scratch. In Catcher, the Benna-Fusi agent sometimes took longer to converge to a good performance in the first few epochs of training, but subsequently became faster than all the control agents in recalling how to perform the task.

We tested the agents' ability to remember over multiple timescales by running the same experiments with different epoch lengths (2500 to 160k) and found that the Benna-Fusi agent demonstrated a better memory than the control in all cases (Figure S1). Furthermore, to ensure that these benefits are not limited to a two-task setting, we ran experiments rotating over three tasks and obtained similar results (Figure S2).

\subsection{Continual Learning within a Single Task}

The continual learning problem is normally posed as the challenge of learning how to perform a series of well-defined tasks in sequence; in RL, however, the issue of nonstationary data often occurs within the training of \textit{one} task. This effect occurs primarily due to (i) strong correlation in time between consecutive states and (ii) changes in the agent's policy altering the distribution of experiences. The most common way to deal with this problem is to use an experience replay database to decorrelate the data, without which the agent can struggle to learn a stable policy (Figure S3). In the final set of experiments, we wanted to see whether using the Benna-Fusi model could enable stable learning in a single task \textit{without} the use of a replay database.

\subsubsection{Experimental Setup}

Control and Benna-Fusi agents were trained on Cart-Pole and Catcher separately in an \textit{online} setting, such that there was no experience replay database and the agents were trained after every time step on the most recent experience.

The architectures of the control and Benna-Fusi agents were the same as in the previous set of experiments bar a couple of differences: the network was smaller (two hidden layers of 100 and 50 units respectively) and, in the Benna-Fusi agent, $g_{1,2}$ was set to a larger value of $0.01$ in order to be able to remember experiences over shorter timescales.

\subsubsection{Results}

\begin{figure}[ht]
\begin{center}
\centerline{\includegraphics[width=\columnwidth]{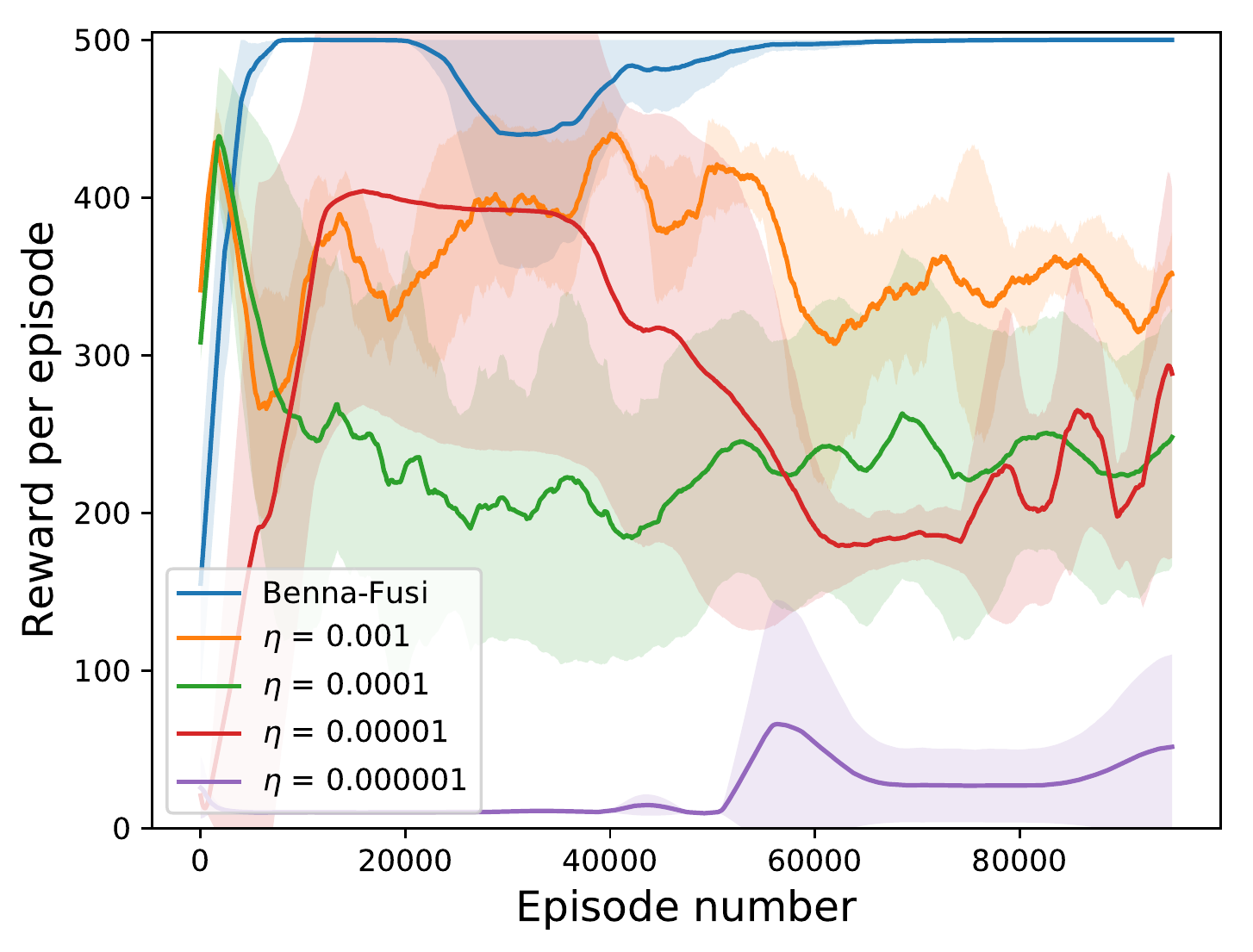}}
\caption{The 1000 test-episode moving average of reward in Cart-Pole for the Benna-Fusi agent and control agents with different learning rates; means and s.d. errorbars over 3 runs per agent.}
\label{cartpole_plots}
\end{center}
\vskip -0.3in
\end{figure}

While none of the control agents were able to learn and maintain a consistently good policy for the Cart-Pole task, the Benna-Fusi agent learned to perform the task to perfection in most cases (Figure \ref{cartpole_plots}). For Catcher, however, all agents were able to learn a consistently good policy, with the control agent learning a bit faster (see Figure S4).

The reason that the control agents struggle to learn a stable policy for Cart-Pole in an online setting, but not Catcher, could be that the distribution of the training data is more nonstationary and thus the agents are more prone to catastrophic forgetting as they learn. A common aspect among control tasks, such as Cart-Pole, is that a successful policy often involves restricting experiences to a small part of the state space \cite{de2016off}. For example, in Cart-Pole the aim is to keep the pole upright, and so if an agent trains for a while on a good policy, it may begin to overwrite knowledge of Q-values in states where the pole is significantly tilted.  Since the agent is constantly learning, it could at some point make an update that causes it to make a wrong action that causes the pole to tilt to an angle that it has not experienced in a while. At this point, the agent might not only perform poorly since it has forgotten the correct policy in this region of the state space, but its policy might be further destabilised by training on these `new' experiences. Furthermore, at this stage the exploration rate might have decayed to a low level, making it harder to relearn. 

One idea is not to let $\epsilon$ to decay to $0$, but in practice we found that this does not solve the problem and can actually make learning less stable (Figure S5). This could be (i) because the agent still overfits to states experienced during a good policy and the extra exploration just serves to perturb it into the negative spiral described above faster than otherwise, or (ii) as noted in \cite{de2016off}, in control tasks the policy often needs to be very fine-tuned in an unstable region of the state space; this requires high-frequency sampling of a good policy and so makes excessive exploration undesirable \cite{de2015importance}. In Cart-Pole, the Benna-Fusi agent succeeds in honing its performance with recent experiences of a good policy while simultaneously remaining robust to perturbations by maintaining a memory of what to do in suboptimal situations that it has not experienced for a while.

In Catcher, a good policy will still visit a large part of the state space and consecutive states are also less correlated in time since fruit falls from random locations at the top of the screen. This may explain why the control agent does not have a problem learning the task successfully.

\section{Related Work}
\label{related_work}

The concept of synaptic consolidation has been applied in a number of recent works that tackle the continual learning problem by adding quadratic terms to the cost function that selectively penalise moving parameters according to how important they are for the recall of previous tasks \cite{ruvolo2013ella,kirkpatrick2017overcoming,zenke2017continual,aljundi2017memory}. In \cite{kirkpatrick2017overcoming}, the importance of each parameter for a task is proportional to its term in the diagonal of the Fisher Information matrix at the end of training. In \cite{zenke2017continual}, the importance factor for a parameter is calculated in an online fashion by determining its contribution to the drop in the loss function over training; in contrast to \cite{kirkpatrick2017overcoming}, which uses a local approximation of importance, this method provides a more global interpretation by considering a parameter's impact over the whole learning trajectory. In \cite{aljundi2017memory}, importance is made proportional to the derivative of the L2-norm of the network output with respect to each parameter; by not relying on the loss function for consolidation, their method can be flexibly used to constrain the model with different data than those that were trained on.

The Benna-Fusi model also constrains parameters to be close to their previous values but, in contrast to the approaches described above, consolidation occurs (i) over a \textit{range} of timescales, (ii) without any derived importance factors, and (iii) without any knowledge of task boundaries. These characteristics are useful for situations where you do not have prior knowledge of when and over what timescale the training data will change, a possibly realistic assumption for robots deployed to act and learn in the real world. Furthermore, the importance factors derived in the other works could feasibly be used to modulate the flow between the hidden variables as a way of combining approaches. 

It must be noted that the idea of modelling plasticity at different timescales to mitigate catastrophic forgetting in a neural network is not new: in \cite{hinton1987using}, each weight is split into separate `fast' and `slow' components, which allows the network to retrieve old memories quickly after training on new data. However, this model was only tested in a very simple setting, matching random binary inputs and outputs, and it is shown in \cite{benna2016computational} that allowing the different components to \textit{interact} with each other theoretically yields much longer memory lifetimes than keeping them separate. The momentum variables in Adam \cite{kingma2014adam} and the soft target updates in \cite{lillicrap2015continuous,polyak1992acceleration} also effectively remember the parameter values at longer timescales, but their memory declines exponentially, i.e. much faster than the power law decay in the Benna-Fusi model.

Other approaches to the continual learning problem that  include the use of networks that grow to incrementally learn new skills \cite{ring1997child,rusu2016progressive}, implicitly training multiple models in one network \cite{goodfellow2013empirical,fernando2017pathnet} and  building generative models to mimic old datasets that are no longer available \cite{shin2017continual}, but these are all orthogonal to the approach used in this paper and could be combined with it.

\section{Conclusion}
In this paper, we took inspiration from a computational model of biological synapses \cite{benna2016computational} to show that expressing each parameter of a tabular or deep RL agent as a dynamical system of interacting variables, rather than just a scalar value, can help to mitigate catastrophic forgetting over multiple timescales.

Our work is intended as a proof of concept that we envisage being extended in several ways. First, the sensitivity of continual learning performance to the parameters of the model, such as the number of hidden variables, should be analysed in order to optimise it and an investigation into the information content held at different depths of the chain could yield more effective readout schemes for the value of each weight. Furthermore, it will be important to test the model's capabilities in a more challenging setting by increasing the number and complexity of tasks, potentially using different architectures such as state-of-the-art actor-critic models \cite{lillicrap2015continuous}, as well as to see if the model can facilitate transfer learning in a series of related tasks. In some initial experiments with larger DQN on tasks from the Arcade Learning Environment \cite{bellemare2013arcade,brockman2016openai}, we found that Benna-Fusi agents struggled to reach the same level of performance as the control agents - the reasons for this will need to be investigated in future work.

Finally, it would be interesting to adapt the model in light of the fact that synaptic consolidation is known to be regulated by neuromodulators such as dopamine, which, for example, has been associated with reward prediction error and exposure to novel stimuli \cite{clopath2008tag}. One could modulate the flow between the hidden variables in the model by factors such as these, or by one of the importance factors cited in the previous section, in order to consolidate memory more selectively and efficiently.

\section*{Acknowledgements}
We would like to thank Tom Schaul for his insightful comments and suggestions.
\bibliography{example_paper}
\bibliographystyle{icml2018}

\begin{appendices}
\onecolumn
\title{Supplementary Material}
\maketitle
\beginsupplement

\section{Experimental details}
Tables of parameters for both the tabular and deep Q-learning experiments are shown below.
\begin{table*}[h]
\caption{Parameter values for Tabular Q-learning experiments}
\label{sample-table}
\begin{center}
\begin{small}
\begin{sc}
\begin{tabular}{lcccr}
\toprule
Parameter & Value \\
\midrule
\# Epochs & 24 \\
\# Episodes/Epoch & 10000 \\
Max \# steps per Episode & 20000 \\
$\gamma$ & 0.9 \\
$\lambda$ & 0.9 \\
$\epsilon$ & 0.05 \\
Learning rate & 0.1 \\
Grid size & 10x10 \\
\# Benna-Fusi variables & 3 \\
Benna-Fusi $g_{1,2}$ & $10^{-5}$ \\
Elig. trace scale factor\textit{*} & 10 \\
\bottomrule
\end{tabular}
\end{sc}
\end{small}
\end{center}
\centering
\textit{*Multiple of eligibility trace that flow between beakers \\ is scaled by in modified Benna-Fusi model}
\end{table*}

\begin{table*}[h]
\caption{Parameter values for Deep RL experiments}
\label{sample-table}
\begin{center}
\begin{small}
\begin{sc}
\begin{tabular}{lcccr}
\toprule
Parameter & Multi-task & Single task \\
\midrule
\# Epochs & 40 & 1 \\
\# Episodes/Epoch & 20000 & 100000 \\
Max \# time steps / episode & 500 & 500 \\
Cart-Pole $\gamma$   & 0.95 & 0.95 \\
Catcher $\gamma$ & 0.99 & 0.99 \\
Initial $\epsilon$ (Epoch start) & 1 & 1 \\
$\epsilon$-decay / episode & 0.9995 & 0.9995 \\
Minimum $\epsilon$ & 0 & 0 \\
Neuron type & ReLU & ReLU \\
Width hidden layer 1 & 400 & 100 \\
Width hidden layer 2 & 200 & 50 \\
Optimiser & Adam & Adam \\
Learning rate & $10^{-3}$ to $10^{-6}$ & $10^{-3}$ to $10^{-6}$ \\
Adam $\beta_1$ & 0.9 & 0.9 \\
Adam $\beta_2$ & 0.999 & 0.999 \\
Experience replay size & 2000 & 1 \\
Replay batch size\textit{*} & 64 & 1 \\
Soft target update $\tau$ & 0.01 & 0.01 \\
Soft Q-learning $\alpha$ & 0.01 & 0.01 \\
\# Benna-Fusi variables & 30 & 30 \\
Benna-Fusi $g_{1,2}$ & 0.001625 & 0.01 \\
Test Frequency (Episodes) & 10 & 10 \\
\bottomrule
\end{tabular}
\end{sc}
\end{small}
\end{center}
\centering
\textit{*Updates were made sequentially as in stochastic \\ gradient descent, not all in one go as a minibatch.}
\vskip -0.1in
\end{table*}
\clearpage
\newpage

\section{Additional Experiments}
\subsection{Varying Epoch Lengths}

\begin{figure}[h]\label{fig:timescale_plot}
\begin{center}
\centerline{\includegraphics[scale=0.3]{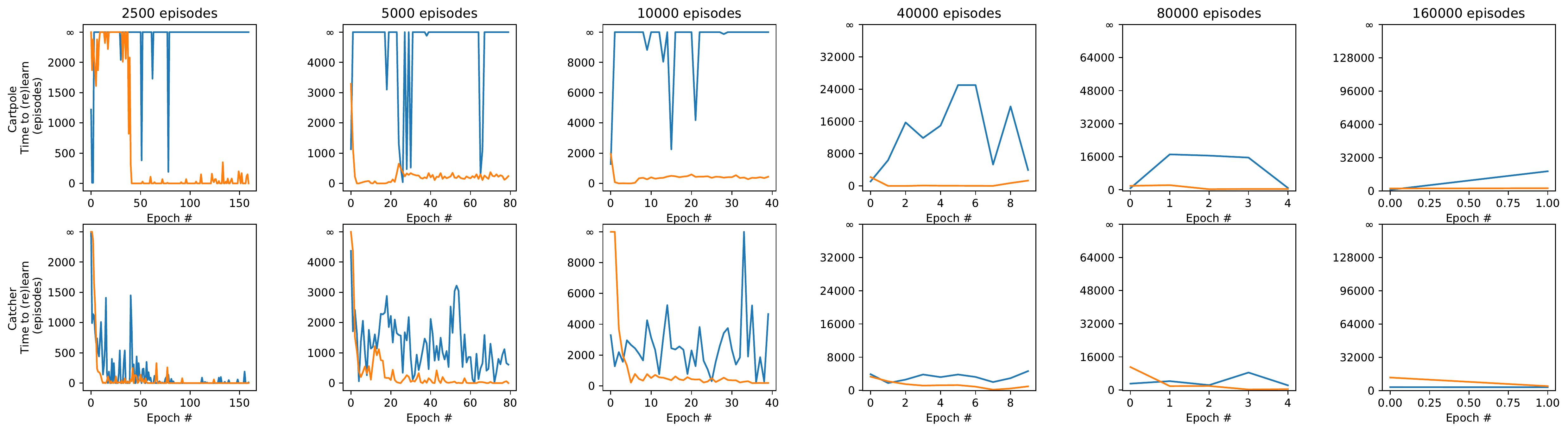}}
\caption{Comparison of time to (re)learn each task in the control agent (blue) and the Benna-Fusi agent (orange) for different epoch lengths. Both agents had a learning rate of 0.001 and the runs with longer epochs were run for fewer epochs. In all cases the Benna-Fusi agent becomes quicker (or in a couple of instances equally quick) at relearning each task than the control agent, demonstrating the Benna-Fusi model's ability to improve memory at a range of timescales.}
\label{timescale_plot}
\end{center}
\vskip -0.2in
\end{figure}

\subsection{Three-task experiments}

In order to ensure that the benefits of the Benna-Fusi model were not limited to the two-task setting, we introduced a new task and ran experiments where training was rotated over the three tasks. The new task was a modified version of Cart-Pole where the length of the pole is doubled (dubbed Cart-PoleLong); our criterion for judging that this task was different enough to Cart-Pole to be considered a new task was that when trained sequentially after Cart-Pole in a control agent, it subsequently led to catastrophic forgetting of its policy for the Cart-Pole task.
\newline

\begin{figure}[h]\label{fig:three_plot}
\begin{center}
\centerline{\includegraphics[width=\textwidth]{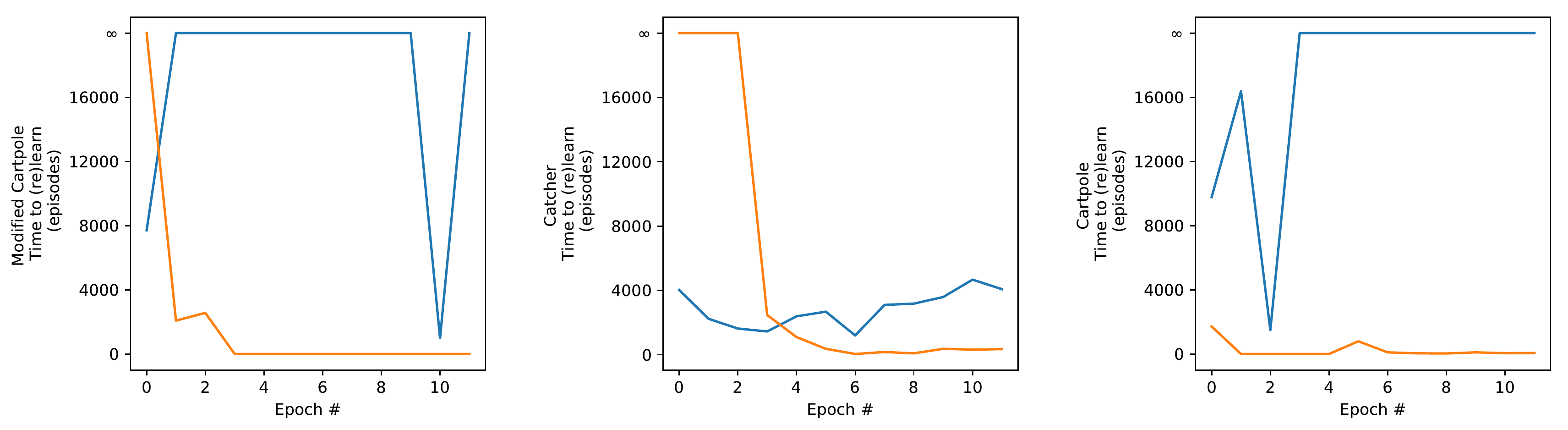}}
\caption{Comparison of time to (re)learn each task in the control agent (blue) and the Benna-Fusi agent (orange) for the three different tasks. Each epoch was run for 20000 episodes and both agents had a learning rate of 0.001. While the Benna-Fusi agent took a little longer to learn Catcher than the control agent, by the end of the simulation the Benna-Fusi agent could learn to recall each task much faster than the control.}
\label{three_plot}
\end{center}
\vskip -0.2in
\end{figure}

Figure \ref{three_plot} shows the remembering times for each task for a control agent and a Benna-Fusi agent when training was rotated over the three tasks (Cart-PoleLong $->$ Catcher $->$ Cart-Pole) over a total of 24 epochs. The results indicate that the Benna-Fusi model exhibits the same benefits as in the two-task setting.

\clearpage
\subsection{Varying size of replay database}
\begin{figure}[h]\label{fig:er_plot}
\begin{center}
\centerline{\includegraphics[scale=0.5]{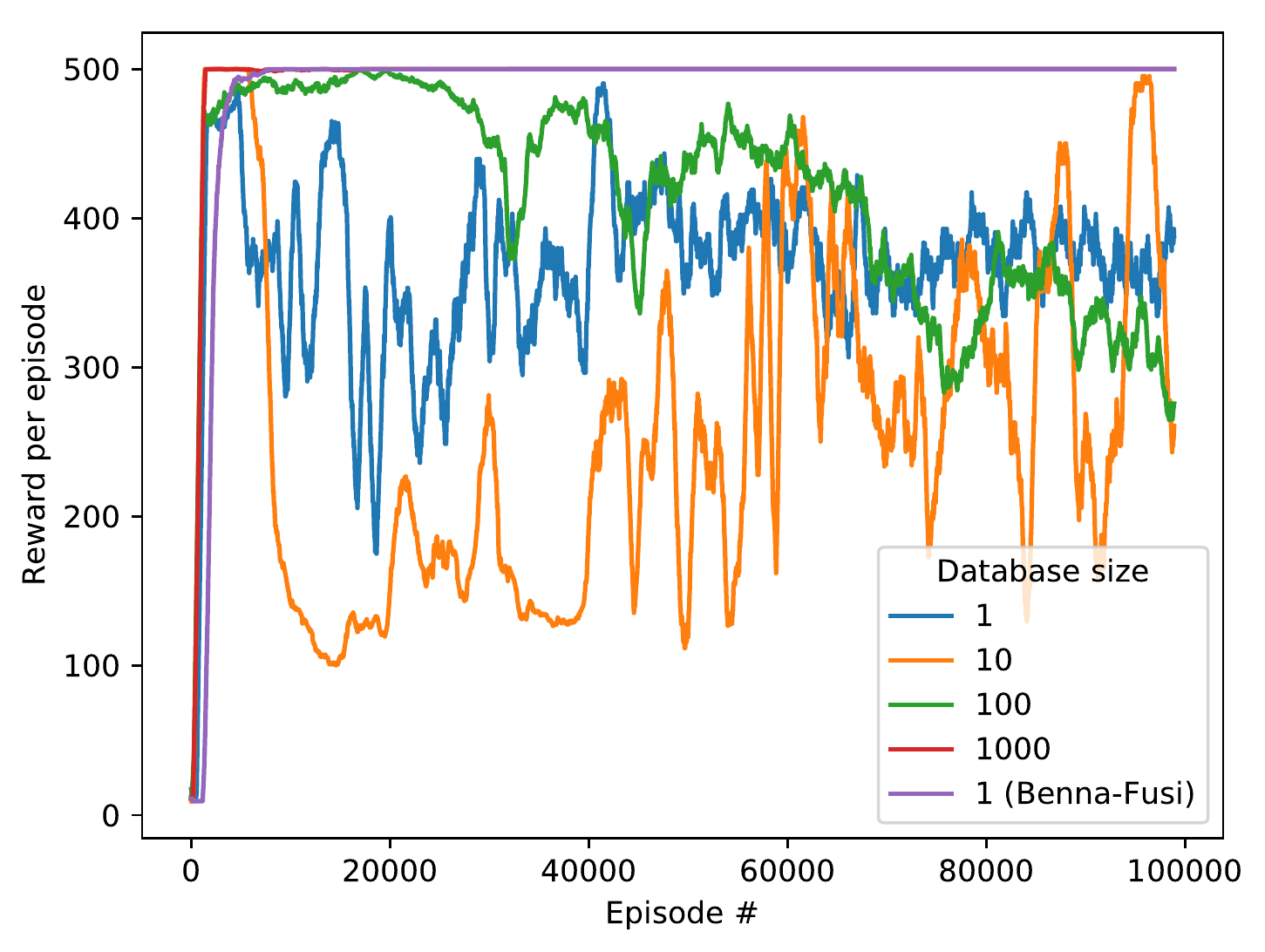}}
\caption{100 test-episode moving average of reward in Cart-Pole for control agents (all with $\eta=0.001$) with different sized experience replay databases and the Benna-Fusi agent in just the online setting. For these experiments, 1 experience was sampled for training from the database after every time step. In the control cases, when the database is too small, the agent can not attain a stable performance on the task while the Benna-Fusi agent can.}
\label{er_plot}
\end{center}
\end{figure}

\subsection{Catcher single task}
\begin{figure}[h]\label{fig:catcher_online}
\begin{center}
\centerline{\includegraphics[scale=0.5]{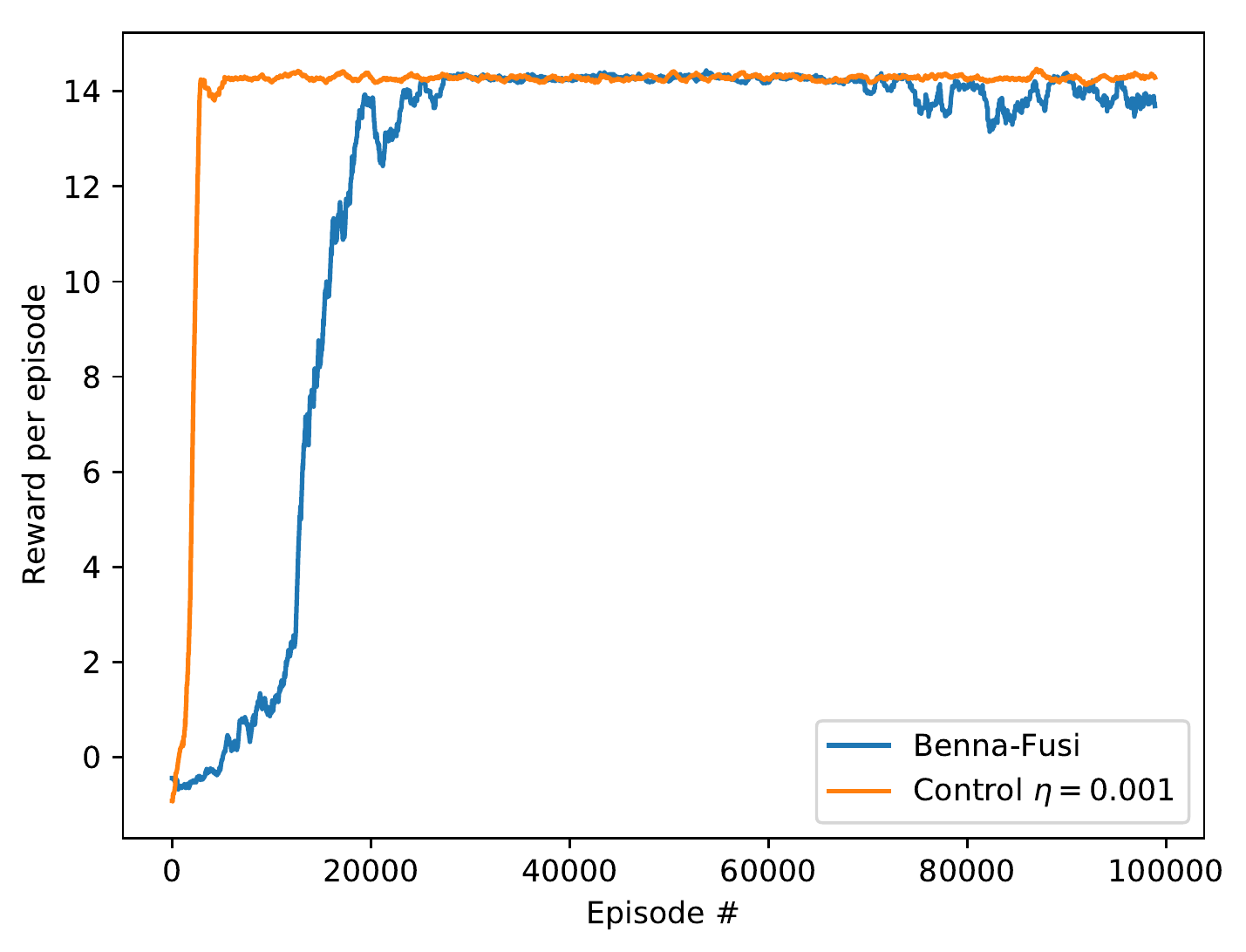}}
\caption{The 100 test-episode moving average of reward per episode in Catcher for the Benna-Fusi agent and the best control agent. The control agent learns faster but both end up learning a good policy.}
\label{catcher_online}
\end{center}
\end{figure}
\newpage 

\subsection{Varying final exploration value}
\begin{figure}[h]\label{fig:min_eps}
\begin{center}
\centerline{\includegraphics[scale=0.5]{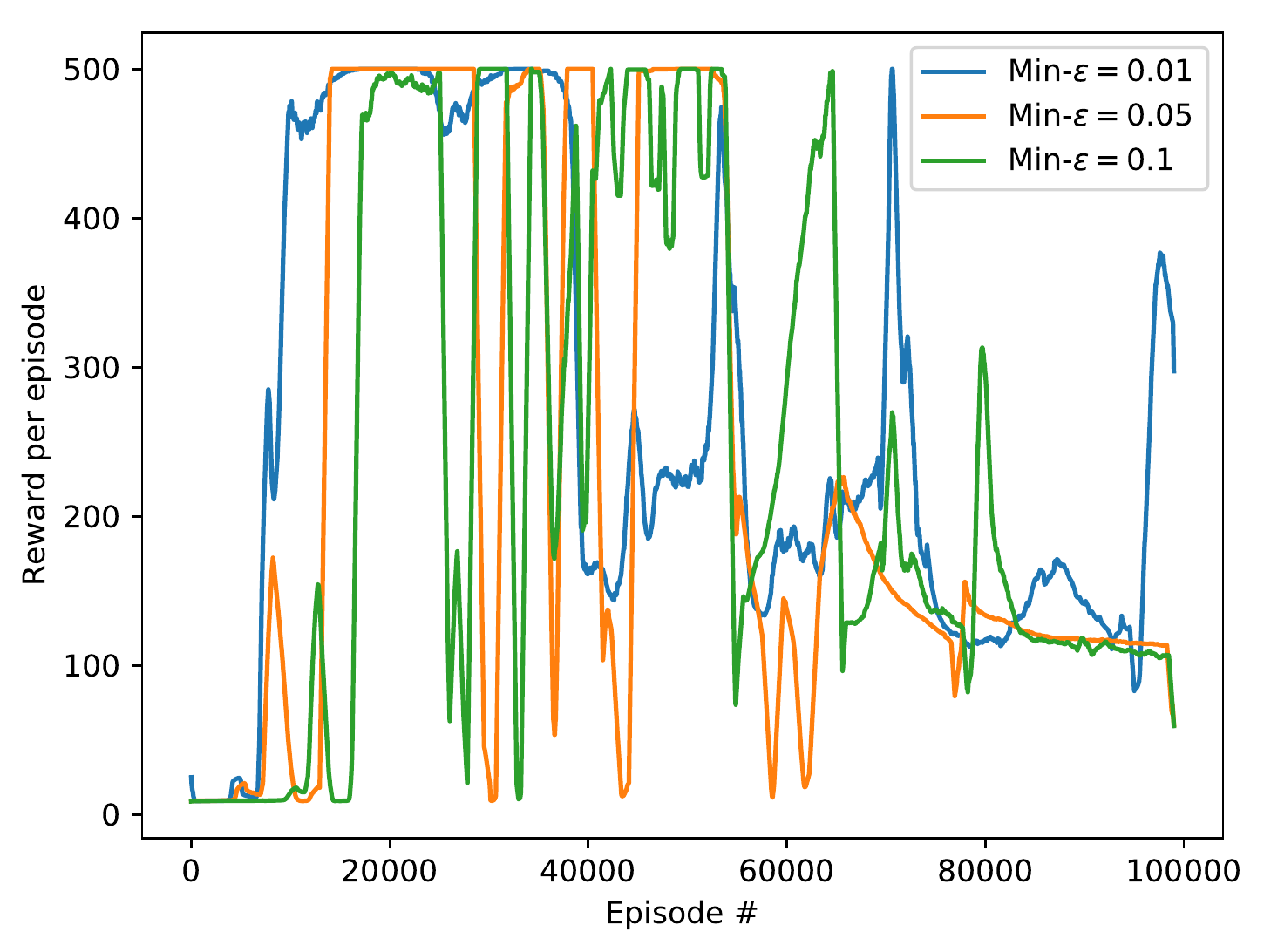}}
\caption{The 100 test-episode moving average of reward per episode in Cart-Pole for control agents where epsilon was not allowed to decay below different minimum values. None of the runs yielded a good stable performance.}
\label{min_eps}
\end{center}
\end{figure}

\end{appendices}
\end{document}